\documentclass[journal]{IEEEtran}
\usepackage{graphicx}
\usepackage{cite}
\usepackage{amssymb}
\usepackage{amsmath}
\usepackage{enumerate}
\usepackage{booktabs}
\usepackage{algorithm}
\usepackage{algorithmic}
\usepackage{listings}
\usepackage[dvipsnames]{xcolor}
\usepackage{color}
\usepackage{colortbl}
\usepackage[colorlinks,linkcolor=blue,anchorcolor=blue,citecolor=green,urlcolor=black]{hyperref}
\usepackage{bbm}
\usepackage{subfigure}
\usepackage{makecell}
\usepackage{soul}

\usepackage{multirow}
\ifCLASSINFOpdf
\else
\fi

\begin{document}
%
\title{RSAdapter: Adapting Multimodal Models for Remote Sensing Visual Question Answering}
%
%
%

\author{Yuduo Wang,
        Pedram Ghamisi,~\IEEEmembership{Senior Member,~IEEE}
\thanks{Y. Wang is with the Department of Computer Science, Humboldt-Universität zu Berlin, 10099 Berlin, Germany, and also with Helmholtz-Zentrum Dresden-Rossendorf, Helmholtz Institute Freiberg for Resource Technology, Machine Learning Group, 09599 Freiberg, Germany. (e-mail: wangyudu@hu-berlin.de).}
\thanks{P. Ghamisi is with Helmholtz-Zentrum Dresden-Rossendorf, Helmholtz Institute Freiberg for Resource Technology, Machine Learning Group, 09599 Freiberg, Germany, and also with Lancaster University, LA1 4YR Lancaster, U.K. (e-mail: p.ghamisi@hzdr.de).}
}

\markboth{IEEE Transactions on Geoscience and Remote Sensing, June 2024}%
{Shell \MakeLowercase{\textit{et al.}}: Bare Demo of IEEEtran.cls for IEEE Journals}

\maketitle

\begin{abstract}
In recent years, with the rapid advancement of transformer models, transformer-based multimodal architectures have found wide application in various downstream tasks, including but not limited to Image Captioning, Visual Question Answering (VQA), and Image-Text Generation. However, contemporary approaches to Remote Sensing (RS) VQA often involve resource-intensive techniques, such as full fine-tuning of large models or the extraction of image-text features from pre-trained multimodal models, followed by modality fusion using decoders. These approaches demand significant computational resources and time, and a considerable number of trainable parameters are introduced. To address these challenges, we introduce a novel method known as RSAdapter, which prioritizes runtime and parameter efficiency. RSAdapter comprises two key components: the Parallel Adapter and an additional linear transformation layer inserted after each fully connected (FC) layer within the Adapter. This approach not only improves adaptation to pre-trained multimodal models but also allows the parameters of the linear transformation layer to be integrated into the preceding FC layers during inference, reducing inference costs. To demonstrate the effectiveness of RSAdapter, we conduct an extensive series of experiments using three distinct RS-VQA datasets and achieve state-of-the-art results on all three datasets. The code for RSAdapter is available online at \href{https://github.com/Y-D-Wang/RSAdapter}{https://github.com/Y-D-Wang/RSAdapter}.
\end{abstract}

\begin{IEEEkeywords}
Remote Sensing, Visual Question Answering, Prameter Effieicent Fine-Tuning (PEFT), Vision-Language Models.
\end{IEEEkeywords}

%
\IEEEpeerreviewmaketitle

\section{Introduction}

\IEEEPARstart{V}{isual Question Answering} (VQA) stands as an interdisciplinary field, situated at the crossroads of computer vision (CV) and natural language processing (NLP). Its primary objective revolves around equipping computational systems with the capacity to comprehend and accurately respond to queries formulated in natural language concerning visual content, encompassing images and videos. Recent years have witnessed substantial advancements in the domain of visual question answering, accompanied by the introduction of essential datasets, evaluation metrics, and increasingly sophisticated models, which have expanded the frontiers of research.

Diverging from conventional visual tasks like image classification \cite{mou2017deep}, object detection \cite{deng2018multi}, and semantic segmentation \cite{xu2022consistency}, the fundamental aim of VQA is to bridge the semantic divide between textual and visual modalities. This objective necessitates the development of models endowed with the capability to reason about the content of visual data based on textual queries. This entails the simultaneous comprehension of both visual content and natural language questions, thereby mandating the integration of computer vision techniques and NLP capabilities. With the advent of the BERT model \cite{devlin2018bert} in 2018, the potential of transformer-based models became apparent. Consequently, transformer-based architectures have gained prominence as a solution for VQA. These architectures typically embark on a two-step process, starting with pretraining on a substantial dataset comprising image-text pairs, followed by fine-tuning on specific VQA datasets to achieve superior performance.

Lu et al. \cite{lu2019vilbert} introduced ViLBERT (Vision-and-Language BERT), a model designed to learn a unified representation of image content and natural language devoid of task-specific biases. Initially, ViLBERT undergoes pretraining on the Conceptual Captions dataset \cite{sharma2018conceptual} and subsequently undergoes fine-tuning for four visual-language tasks: visual question answering, visual commonsense reasoning, referring expressions, and caption-based image retrieval. ViLBERT's architecture requires only minimal modifications when employed for downstream tasks. Subsequently, similar approaches have been presented by other researchers \cite{li2019visualbert, tan2019lxmert}, all involving the fine-tuning of pre-trained models for images and language separately before integrating them. However, this approach has encountered challenges concerning the harmonization of visual-linguistic aspects. The crux of the task lies in the aggregation of multimodal information.

To attain a more universal representation, Su et al. \cite{su2019vl} introduced VL-BERT, which was pre-trained on a comprehensive concept annotation dataset and textual corpora. Chen et al. \cite{chen2020uniter} proposed UNITER, a universal image-text representation method achieved through extensive pretraining on four diverse image-text datasets, including COCO \cite{lin2014microsoft}, Visual Genome (VG) \cite{krishna2017visual}, Conceptual Captions \cite{sharma2018conceptual}, and SBU Captions \cite{ordonez2011im2text}. UNITER offers support for a wide array of heterogeneous downstream tasks by jointly embedding multimodal information.

VQA has recently gained significant attention in the field of Earth Science and Remote Sensing (RS), particularly with the introduction of several new datasets \cite{lobry2020rsvqa, zheng2021mutual, lobry2021rsvqa}. This makes it a current and cutting-edge research topic in these domains. Lobry et al. \cite{lobry2020rsvqa} were among the first to introduce a simple method where features were extracted separately from images using CNN and from text using LSTM. These features were then combined through element-wise multiplication to obtain a joint representation. Subsequent works \cite{yuan2022easy, zhang2023spatial} in this field explored the use of visual features at different levels to enhance the understanding of image content, thereby improving the model's ability to learn from both images and text. The methods mentioned above are all based on traditional CNN and RNN approaches. Similarly, models based on transformers have also been proposed in this context. Bazi et al. \cite{bazi2022bi} utilized the CLIP \cite{radford2021learning} model to extract image and text features. These features were then separately fed into two parallel decoder transformers to learn a joint representation. Siebert et al. \cite{siebert2022multi} jointly fed processed image and text features into VisualBert \cite{li2019visualbert} for better understanding of image-text features. Hackel et al. \cite{hackel2023lit} presented a lightweight transformer-based VQA model in remote sensing and also published a PyTorch-based library for rapid development of Image-Language Models.

Without a doubt, the groundbreaking work mentioned above has greatly advanced the research in RS-VQA. However, existing transformer-based methods still exhibit at least three limitations:

\begin{itemize}
    \item In the domain of natural images, it has been demonstrated that well pre-trained multimodal models can offer strong transferability to downstream VQA tasks through fine-tuning. RS images typically exhibit complex backgrounds, high similarity between different objects, and large object size variations, making them more challenging to process than natural images. Therefore, there are doubts on whether pretrained transformer models used for natural images still exhibit strong transferability on RS-VQA tasks.
    \item The majority of methods require updating all parameters of the pre-trained model to adapt to the RS-VQA tasks and achieve satisfactory performance. As the scale of pre-trained large models rapidly expands, for example, the Llama2 model \cite{touvron2023llama} with 70B parameters,
    saving computational costs has remained a significant challenge.
    \item There is still debate over whether it is necessary to perform full fine-tuning on pre-trained multimodal models, given their demonstrated strong transferability to downstream tasks. Directly applying full fine-tuning techniques of pre-tranined large models on data-limited RS-VQA tasks may lead to overfitting, which results in inferior performance, instability, and reduced generalization ability. This, in turn, affects the effectiveness of utilizing pre-trained models.
\end{itemize}

To overcome the limitations mentioned above, we propose a novel approach, namely RSAdapter, to efficiently fine-tune pre-trained multimodal models, enabling them to better adapt to RS-VQA tasks. Specifically, we start with the ViLT model \cite{kim2021vilt} as our backbone, and then we compare the performance differences between inserting RSAdapter in parallel next to the multiheaded self attention (MSA) and feedforward network (MLP) components separately. The results consistently show that RSAdapter inserted next to the MLP component performed better. Finally, we simultaneously insert RSAdapter next to MSA and MLP components while adding corresponding scaling layers to control the contribution of each RSAdapter to the model effectively. To validate the effectiveness of RSAdapter, we conduct comprehensive experiments on three different RS-VQA datasets, and the results indicated that when freezing all parameters of the pre-trained multimodal model and only updating the newly added lightweight RSAdapter, our proposed method achieves better results than previous state-of-the-art methods. The main contributions of this paper are summarized as follows:
\begin{enumerate}
    \item We propose a streamlined architecture that leverages vision and language models, with a particular focus on runtime and parameter efficiency.
    \item In contrast to conventional methods for RS-VQA, our approach achieves impressive performance without the reliance on region-based features or complex feature extractors for visual and textual embeddings. Moreover, we introduce a novel re-parameterization method, RSAdapter, which can achieve better performance without increasing the number of parameters in the inference phase. Compared to RSVQA, our inference time can be reduced by 43\% on the RS-LR dataset.
    \item We conduct a comprehensive set of experiments using three distinct RS-VQA datasets to validate the effectiveness of our proposed method. Particularly noteworthy is its ability to achieve competitive performance, even when training data is limited.
\end{enumerate}

The remainder of this paper is structured as follows: Section II provides a review of the related work pertinent to this study. Section III offers a comprehensive description of the proposed method. Section IV presents details about the datasets utilized in this study and outlines the experimental results. Conclusions and additional discussions are consolidated in Section V.

\section{Related Work}

\subsection{Vision-Language Transformer Models}

Pre-trained Vision-Language models have demonstrated impressive performance on downstream multimodal tasks, including Image Captioning \cite{chang2023changes}, Visual Question Answering \cite{antol2015vqa}, image-text generation \cite{xu2022txt2img}, and various NLP and CV tasks. For vision-language transformer models, they are typically categorized into two distinct classes known as single-stream transformers and dual-streams transformers.

Currently, most vision-language models use a single-stream approach, which includes only one transformer stack together. They concatenate visual tokens and textual tokens as the input to the transformer. It significantly simplifies the architecture compared to the  dual-streams transformers, achieving comparable performance with fewer parameters. Additionally, it offers ease of scalability, as we can handle other modalities' questions, such as video and audio, by simply concatenating tokens from other modalities. ViLT \cite{kim2021vilt}, PixelBERT \cite{huang2020pixel}, VisualBERT \cite{li2019visualbert} all belong to the category of single-stream transformers. They all use BERT \cite{devlin2018bert} to extract text embeddings, and both PixelBERT and VisualBERT employ CNNs as feature extractors for the image modality. After specific processing, the image features are concatenated with text tokens and fed into the transformer. What sets ViLT apart from other models is its use of a straightforward linear projection to obtain image embeddings. This simplifies the structure of ViLT, leading to increased runtime and parameter efficiency. It is precisely this characteristic that led us to choose ViLT as our backbone model in this paper.

\subsection{Parameter-efficient Fine-tuning}

Recent research has underscored the significant potential of large-scale models. The ever-increasing model sizes have, in turn, escalated the demand for computational resources required to fine-tune these models for downstream tasks. This trend has sparked substantial research interest in the realm of Parameter-efficient Fine-tuning (PEFT). Initially, PEFT methods found their applications in Natural Language Processing (NLP), offering efficient transfer of large pre-trained models to downstream tasks with the introduction of only a small number of trainable parameters, while keeping the original large model parameters frozen. In certain tasks, this approach can even outperform the performance achieved by fully fine-tuning the model.

The pioneering PEFT method, Adapter \cite{houlsby2019parameter}, as proposed by Houlsby et al., involves embedding a compact task-specific MLP network into a large pre-trained model to facilitate adaptation to downstream tasks. Prefix tuning \cite{li2021prefix} and Prompt tuning \cite{lester2021power} entail the fine-tuning of large pre-trained models through the introduction of new trainable task-specific vectors or virtual token embeddings. Low-Rank Adaptation \cite{hu2021lora} (LoRA) augments the parameters by introducing trainable low-rank matrices to the multi-head attentions in the large pre-trained model. Building upon these foundational PEFT methods, subsequent works \cite{mao2022unipelt, he2021towards, luo2023towards} have achieved superior results on downstream tasks by combining various techniques. In addition to the aforementioned mainstream methods, Bitfit \cite{zaken2022bitfit} has demonstrated that updating only the bias terms can yield satisfactory results in specific downstream tasks.

RSAdapter adopts a design akin to Adapter \cite{houlsby2019parameter} but enhances it with re-parameterization tailored specifically for the linear layer. Unlike traditional adapter methods, RSAdapter yields performance improvements without introducing additional parameters during the inference stage.

\subsection{Remote Sensing VQA}

Lobry et al. \cite{lobry2020rsvqa} were the first to introduce VQA to the RS field. Their work initially released two new datasets, referred to as RS-VQA Low Resolution and RS-VQA High Resolution. Subsequently, they proposed a simple joint approach, which involved extracting features from both images and text data using CNN and LSTM separately, followed by computing the dot product of image-text features and feeding them into an MLP for answer prediction. Later, they released a large-scale VQA dataset \cite{lobry2021rsvqa} dedicated to remote sensing, containing close to 15 million samples.

In the subsequent work by Zheng et al. \cite{zheng2021mutual}, another dataset, RS-IVQA, was introduced. This dataset was generated based on several existing datasets in the RS domain, including \cite{xia2017aid, yang2010bag, zhang2014saliency, xia2018dota, zhang2019hierarchical}, to create relevant image-question pairs. This work also introduced the MAIN method, which leverages attention mechanisms and bilinear techniques to enhance the relationship between spatial positions and textual information. Building upon the three aforementioned datasets, Yuan et al. \cite{yuan2022easy} proposed a text-guided multi-level visual feature learning approach. Additionally, the SPCL method based on difficulty was introduced in the final answer prediction stage.

In contrast to the methods mentioned above that use traditional CNN and LSTM models for feature extraction, Bazi et al. \cite{bazi2022bi} introduced a transformer-based VQA approach. They initially utilized the CLIP \cite{radford2021learning} model to separately extract text and image features, and then employed transformer decoders based on co-attention to capture the relationship between images and text. Chappuis et al. \cite{chappuis2022prompt} initially processed image information for classification and generated text prompts. These prompts were then input into a language model for answer prediction. Siebert et al. \cite{siebert2022multi} employed the VisualBERT \cite{li2019visualbert} model to better learn joint representation.

Recently, Zhang et al. \cite{zhang2023spatial} introduced a hash-based spatial multiscale visual learning method to enhance the perception of spatial positional information. Additionally, in the final visual-text fusion stage, they employed a complex interaction module to capture image-text interaction information.

It is worth noting that Bazi et al. \cite{bazi2022bi} introduced an encoder-decoder structure, increasing the model's complexity, while Siebert et al. \cite{siebert2022multi} performed full fine-tuning on VisualBERT \cite{li2019visualbert}, requiring substantial computational resources and runtime. Compared to these existing transformer-based methods, RSAdapter achieves efficient fine-tuning without increasing model complexity, saving training time and computational resources. Based on these points, we believe that the proposed method is an effective addition to the current RS-VQA field.

\section{Methodology}

\begin{figure*}
    \centering
    \includegraphics[width=\textwidth]{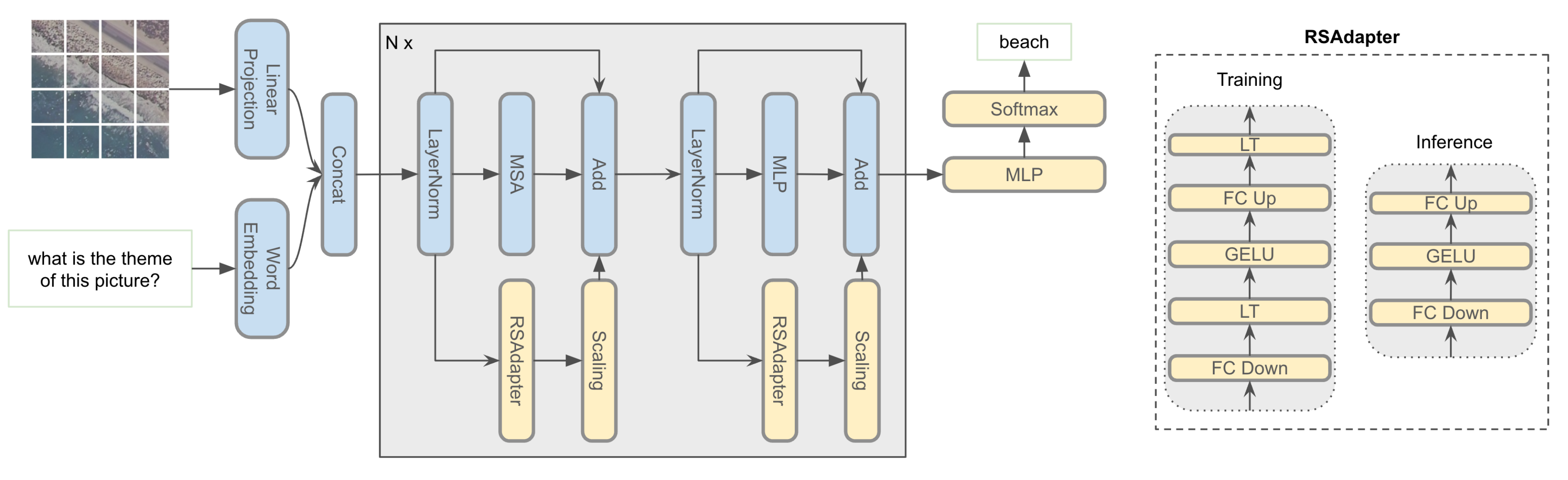}
    \caption{Graphical illustration of the proposed RSAdapter. We insert the RSAdapter next to the MSA and  MLP in the transformer block. Among them, the blue block is in the frozen state during training, while the yellow block will be updated. During inference, the weights and biases in the linear transformation can be merged into the preceding fully connected (FC) layer. LT indicates linear transformation.}
    \label{fig:1}
\end{figure*}

In this section, we first describe our baseline model, ViLT \cite{kim2021vilt}. Then, we introduce adapter, RSAdapter, and scaling RSAdapter to show how we adapt a pre-trained multimodal model for effective RS-VQA step by step. As shown in Fig. \ref{fig:1}, the weights and biases in the linear transformation of RSAdapater can be merged into the preceding fully connected (FC) layer to reduce the inference cost.

\subsection{Preliminary}

After the introduction of the Vision Transformer (ViT) by Dosovitskiy et al. \cite{dosovitskiy2020image}, transformers have made significant inroads into the field of CV. Transformer-based pre-trained language models (PLMs) have consequently found widespread application in a variety of multi-modal tasks, including VQA \cite{antol2015vqa}. In this study, our objective is to design an efficient network based on a pre-trained multimodal model for the purpose of RS-VQA. Additionally, we aim to assess its performance in comparison to other existing models that have been specifically tailored for this task.

In a typical ViT model, an image is traditionally divided into a series of small patches. To maintain the original ViT model methodology as closely as possible, while avoiding the introduction of additional complex modules, we have selected ViLT \cite{kim2021vilt} as our baseline model. Specifically, ViLT utilizes a pre-trained BERT \cite{devlin2018bert} to extract word and position embeddings for the text component. For images, it employs patch projection instead of heavy feature extraction models to extract regional or grid features. 

Given input text $t \in \mathbb{R}^{L}$ and input image $x \in \mathbb{R}^{H \times W \times C}$, ViLT first embeds the text into $\Bar{t} \in \mathbb{R}^{n_t \times d}$ via BERT \cite{devlin2018bert}. Then, ViLT utilizes linear projection \cite{dosovitskiy2020image} to transform image $x$ into visual tokens $\Bar{v} \in \mathbb{R}^{n_v \times d}$. Two learnable tokens, denoted as $t_{class} \in \mathbb{R}^{1\times d}$ and $v_{class} \in \mathbb{R}^{1\times d}$, are concatenated with $\Bar{t}$ and $\Bar{v}$, respectively. This can be expressed as:
\begin{equation}
    T = [t_{class}, \Bar{t}_1, \dots, \Bar{t}_{n_t}] + P_t
\end{equation}
\begin{equation}
    V = [v_{class}, \Bar{v}_1, \dots, \Bar{v}_{n_v}] + P_v
\end{equation}
where $P_t \in \mathbb{R}^{(n_t + 1) \times d}$ and $P_v \in \mathbb{R}^{(n_v + 1) \times d}$ are the positional embeddings for text embeddings and image embeddings. To enhance the model's ability to discern the relationship between images and text, corresponding modal-type embeddings are added to text and image embeddings. Text modal-type embedding $t_{type}$ is assigned $0$, and image modal-type embedding $v_{type}$ is assigned $1$, enabling the model to differentiate between images and text. They are then concatenated to create a unified token denoted as $X_0$, which can be formulated by,
\begin{equation}
    X_0 = [T + t_{type}, V + v_{type}].
\end{equation}
This combined token serves as the ultimate input embedding for a series of transformer blocks. The calculation within a standard transformer block can be represented as follows:
\begin{equation}
    X_{l}^{\prime} = X_{l-1} + \text{MSA}(\text{LN}(X_{l-1}))
\end{equation}
\begin{equation}
    X_{l} = X_{l}^{\prime} + \text{MLP}(\text{LN}(X_{l}^{\prime}))
\end{equation}
where $X_{l-1}$ and $X_l$ denote the input and output of the $l$-th transformer block, MSA, MLP and LN denote the multiheaded self attention, feedforward network, and layer normalization, respectively. In particular, MSA can be defined as
\begin{equation}
    \text{MSA}(X) = \text{Concat(Head$_1 (X)$, $\dots$, Head$_h (X)$)}W^O \\
\end{equation}
\begin{equation}
    \text{Head$_i$}(X) = \text{Softmax}(\frac{(XW_{i}^{Q})(XW_{i}^{K^\mathrm{T}})}{\sqrt{d_k}} (XW_{i}^{V}))
\end{equation}
where $\text{Head$_i$}(X)$, $W_{i}^{Q} \in \mathbb{R}^{d \times \frac{d}{h}}$, $W_{i}^{K} \in \mathbb{R}^{d \times \frac{d}{h}}$ and $W_{i}^{V} \in \mathbb{R}^{d \times \frac{d}{h}}$ are the $i$-th head dot-product attention and weight matrices. $W^O \in \mathbb{R}^{d \times d}$ is the weight matrix for the output transformation. MLP can be formulated by
\begin{equation}
    \text{MLP}(X) = f(X W^{m_1} + b_{m_1})W^{m_2} + b_{m_2}
\end{equation}
Here, $W^{m_1} \in \mathbb{R}^{d\times 4d}$, $W^{m_2} \in \mathbb{R}^{4d\times d}$ and $b_{m_1} \in \mathbb{R}^{4d}$, $b_{m_2} \in \mathbb{R}^{d}$ represent the projection weights and biases. $f(\cdot)$ is a non-linear activation function, typically the GELU function \cite{devlin2018bert}.

\subsection{Multimodal Adapter}

The pre-trained multimodal model has already demonstrated excellent results in VQA tasks \cite{li2019visualbert, kim2021vilt, lu2019vilbert}. Recently, multimodal models \cite{alayrac2022flamingo, guo2023images} have even exhibited remarkable few-shot and zero-shot capabilities because they effectively learned the relationships between image-text pairs. Despite our dataset consisting of RS images, which significantly differ from the natural images in the pre-trained dataset, our results indicate that we can still achieve competitive outcomes through efficient fine-tuning. This is possible because the multimodal model has already undergone effective pre-training on extensive image-text datasets, giving it strong transferability. Inspired by efficient fine-tuning techniques in NLP and CV, we have adopted the Adapter framework \cite{houlsby2019parameter}.

To enable efficient fine-tuning, the transformer layer incorporates an Adapter. The adapter module acts as a bottleneck, providing a limited set of learnable parameters. It consists of the dimension reduction for decreasing feature dimensions, the application of a non-linear activation function (typically the GELU function \cite{devlin2018bert}), and the dimension expansion for restoring the original dimensions. Therefore, given the input $X \in \mathbb{R}^{n \times d}$, the output is computed as follows:
\begin{equation}
    X^{\prime} = f(X W^{down})W^{up} + X
    \label{equation:1}
\end{equation}
where $W^{down} \in \mathbb{R}^{d \times d^{\prime}}$ and $W^{up} \in \mathbb{R}^{d^{\prime} \times d}$ represent the down-sampling matrix and up-sampling matrix, respectively. $f(\cdot)$ is the GELU function.

In typical scenarios, the Adapter can be inserted into two different positions. The first approach involves sequentially inserting the Adapter after MSA. In this case, the Eq.~(\ref{equation:1}) can be modified as follows:
\begin{equation}
    X^{\prime} = f(\text{MSA($X$)} W^{down})W^{up} + X
\end{equation}

The second method involves inserting the Adapter sequentially after MLP. In this case, the Eq.~(\ref{equation:1}) can be written as follows:
\begin{equation}
    X^{\prime} = f(\text{MLP($X$)} W^{down})W^{up} + X.
\end{equation}

The methods described above have already proven to be effective in various tasks \cite{houlsby2019parameter}, but they were primarily designed for NLP tasks. To adapt the pre-trained visual-text features to remote sensing data, we insert the Adapter in parallel and remove the skip connection. In this case, given the transformer block input $x \in \mathbb{R}^{n \times d}$, where $n$ is the length of tokens and $d$ is the transformer feature size, the output is calculated by
\begin{equation}
    X^{\prime} = g(X) + f(X W^{down})W^{up} \label{equation:2}
\end{equation}
where $g(\cdot)$ represent either the MSA  or MLP operation. This parallel approach allows the Adapter to be more flexibly inserted into various parts of the transformer block without altering the overall structure of the transformer. To maintain consistency with the original methods as much as possible, our primary focus has been on studying the parallel insertion of the Adapter alongside MSA and MLP.

\subsection{RSAdapter}

\begin{figure}
    \centering
    \includegraphics[width=0.75\linewidth]{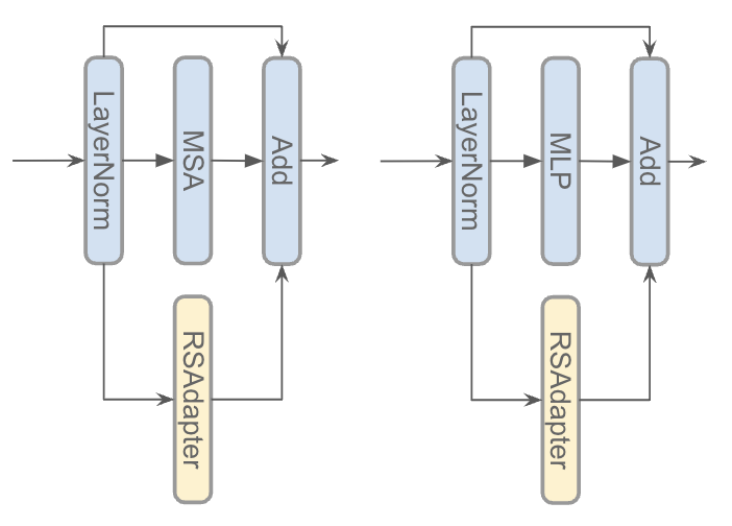}
    \caption{Two possible insert positions for RSAdapter.}
    \label{fig:2}
\end{figure}

Re-parameterization techniques have been widely applied in the field of CV \cite{ding2019acnet, ding2021repvgg}. Here, we introduce a novel re-parameterization method that allows us to achieve better performance during inference without incurring additional costs, building upon the foundation of the parallel Adapter. We first add a linear transformation after each Linear layer in the parallel Adapter. In this case, the Eq.  (\ref{equation:2}) can be rewritten as
\begin{equation}
    X^{\prime} = g(X) + \phi_{up} (f(\phi_{down} (X W^{down}))W^{up})
    \label{equation:3}
\end{equation}
where $\phi_{up}$ and $\phi_{down}$ are the linear transformations applied after the respective linear layers. Regarding the right half of the Eq. (\ref{equation:3}), we can re-parameterize it for inference by
\begin{equation}
\begin{split}
    \phi_{rep}(X) &= (XW + b)W^{\prime} + b^{\prime}, \\ 
    &= XWW^{\prime} + bW^{\prime} + b^{\prime}, \\
    &= XW_{rep} + b_{rep}.
    \label{equation:4}
\end{split}
\end{equation}
Here, $W \in \mathbb{R}^{d \times d^{\prime}}$ and $W^{\prime} \in \mathbb{R}^{d^{\prime} \times d^{\prime}}$ are the down or up-projection weight matrix and linear transformation weight matrix respectively. $W_{rep} = WW^{\prime}$ and $b_{rep} = bW^{\prime} + b^{\prime}$ are the weight and bias that have been re-parameterized. Therefore, based on Eq.(\ref{equation:4}), Eq.(\ref{equation:3}) during the inference phase can be rewritten as follows:

\begin{equation}
    X^{\prime} = g(X) + \phi_{up}^{rep} (f(\phi_{down}^{rep}(X)))
\end{equation}
where $\phi_{up}^{rep}(\cdot)$ and $\phi_{down}^{rep}(\cdot)$ are the corresponding re-parametrize function. Fig. \ref{fig:2} illustrates the insertion of the RSAdapter into the respective MSA and MLP.

\subsection{Scaling RSAdapter}

After introducing the MSA RSAdapter and MLP RSAdapter to better control the impact of the RSAdapter on the corresponding MSA and MLP, we add a scaling layer after each RSAdapter. To maintain structural simplicity, we implement the scaling layer for different positions of the RSAdapter using a unique scaling factor. Eq.(\ref{equation:3}) can be modified by:
\begin{equation}
    X^{\prime} = g(X) + s \cdot \phi_{up} (f(\phi_{down} (X W^{down}))W^{up})
\end{equation}

The final structure of the adopted transformer block is shown in Fig. \ref{fig:1} and the PyTorch style pseudocode of the adopted ViLT layer is shown in Algorithm \ref{alg:1}. The computation of the adopted block can be written as:
\begin{equation}
\begin{split}
    X_{l}^{\prime} & = s_{a} \cdot \phi_{up} (f(\phi_{down} (\text{LN}(X_{l-1}) W^{down}))W^{up}) \\
    &+ \text{MSA}(\text{LN}(X_{l-1})) + X_{l-1}  
\end{split}
\end{equation}
\begin{equation}
\begin{split}
    X_{l} & = s_{p} \cdot \phi_{up} (f(\phi_{down} (\text{LN}(X_{l}^{\prime}) W^{down}))W^{up}) \\
    &+ \text{MLP}(\text{LN}(X_{l}^{\prime})) + X_{l}^{\prime}
\end{split}
\end{equation}
where $X_{l-1}$, $X_{l}^{\prime}$ and $X_{l}$ denote the input, intermediate output, and final output of the $l$-th transformer block. Here, $s_{a}$ and $s_{p}$ are scaling factors for the corresponding RSAdapters.

\begin{algorithm}[H]
\caption{Pseudocode of the Adopted ViLT Layer}
\label{alg:1}
\begin{lstlisting}[
 language=Python,
 basicstyle=\ttfamily,
 breaklines=true
] 
def viltlayer_with_rsadapter(x):
    x_ln = LN(x)
    x_attn = Attention(x_ln)
    attn_ada = RSAdapter(x_ln)
    x = x_attn + s_a * attn_ada + x
    x_ln = LN(x)
    mlp_out = MLP(x_ln)
    mlp_ada = RSAdapter(x_ln)
    x = mlp_out + s_p * mlp_ada + x

    return x
\end{lstlisting}
\end{algorithm}

\subsection{Predicting Answers}

For the final prediction, we simply take the first [class] token of the last transformer block and feed it to the classification head. Give $y \in \mathbb{R}^{d}$, we feed this token to a 3-layer MLP with the softmax activation function. We formulate the problem as a classification task, in which each possible answer is a class. Therefore, the size of the output vector depends on the number of possible answers. The answer class probability $p_y \in \mathbb{R}^{\#class}$ can be written as follows:

\begin{equation}
    p_{y} = \text{Softmax}(\text{MLP}(y))
\end{equation}

\noindent During the training process, we employ standard cross-entropy loss as our optimization objective to train the network.

\section{Experiments}
\subsection{Dataset Descriptions}
We adopt three benchmark datasets to evaluate the performance of the proposed method on RS-VQA tasks.
\subsubsection{\textbf{RS-VQA Low Resolution}}

This dataset \cite{lobry2020rsvqa} comprises a total of 772 images, each with dimensions of 256 x 256 pixels, captured with a spatial resolution of 10 meters. These images collectively cover an area of 6.55 square kilometers and were obtained through the Sentinel-2 satellite over the airspace of the Netherlands. The dataset encompasses a grand total of 77,232 image-question-answer triplets, with each image associated with approximately 100 different questions and their corresponding answers. All the questions are categorized into four distinct types: count, presence, comparison and rural/urban. Following the recommendation from \cite{lobry2020rsvqa}, we have established a total of nine possible answers. We use the pre-divided portions of the dataset, with 77.8\% serving as the training set, 11.1\% as the validation set, and the remaining 11.1\% as the test set to evaluate the model's effectiveness.

\subsubsection{\textbf{RS-VQA High Resolution}}

The dataset \cite{lobry2020rsvqa} comprises a total of 10,659 images, each with dimensions of 512 x 512 pixels and a spatial resolution of 15 cm, covering a total area of 5,898 square meters. All the images were extracted from the USGS High-Resolution Orthoimagery (HRO) dataset. In total, the dataset includes 1,066,316 image-question-answer triplets, with approximately 100 questions and their corresponding answers for each image. All the questions are categorized into four different types: count, presence, comparison, and area. After a thorough analysis of all the answers, we have identified a total of 94 possible answers. For our experiments, we use 61.5\% of the dataset as the training set, 11.2\% as the validation set, and the remaining 27.3\% as the test set to evaluate our model's effectiveness. It is important to note that this dataset includes two separate test sets, HR1 and HR2. Among these test sets, HR1 covers an area similar to the training and validation sets, while HR2, designed to assess the model's generalization capabilities, covers an area entirely distinct from the training set. Additionally, HR2 uses data collected by a different sensor.

\subsubsection{\textbf{RS-IVQA}}

The dataset \cite{zheng2021mutual} primarily focuses on two fundamental aspects of remote sensing: scene classification and object detection. The dataset comprises a total of 37,264 images, primarily sourced from existing remote sensing datasets such as AID \cite{xia2017aid}, UC-Merced (UCM) \cite{yang2010bag}, Sydney \cite{zhang2014saliency}, DOTA \cite{xia2018dota} and HRRSD \cite{zhang2019hierarchical}. In total, the dataset includes 111,134 image-question-answer triples, all questions were categorized into three main types: yes/no, number and others. After a comprehensive analysis of all the answers, we have established a total of 519 possible answer categories. To ensure a more accurate comparison of the models' effectiveness, we adopt the same dataset split as \cite{zheng2021mutual}, utilizing stratified sampling to allocate 80\% of the dataset to the training set, 10\% to the validation set, and the remaining 10\% to the test set.

\subsection{Experimental Settings and Implementation Details}

In our experiment, we use ViLT${B32}$ as our baseline model \cite{kim2021vilt}, with $d$ set to 768 and the number of transformer layers $N$ set to 12. We implement our model using the Hugging Face Transformers library \cite{wolf2020transformers}. The default size of $d^{\prime}$ is set to 192, and GELU is used as the non-linear activation function. The weights $W^{\prime}$ and biases $b^{\prime}$ are initialized to ones and zeros, respectively, for both the $\phi_{up}$ and $\phi_{down}$ linear transformations. Similarly, $s_{a}$ and $s_{p}$ are initialized to 1.

In the training phase, we set the maximum number of iterations to 50 epochs for the LR and RSIVQA datasets and 20 epochs for the HR dataset. We use a batch size of 64. For the LR and RSIVQA datasets, we employ a learning rate of 1e$^{-3}$ for the first 4 epochs to warm up our model, after which we optimize the model using a learning rate of 1e$^{-5}$. In the case of the HR dataset, the warm-up learning rate is set to 1e$^{-4}$, and the normal learning rate is 1e$^{-6}$. We utilize the Adam optimizer \cite{kingma2014adam} for model optimization.

We employ Randaugment \cite{cubuk2020randaugment} for data augmentation in the LR and HR datasets. For LR and HR datasets, we use original image sizes of 256 $\times$ 256 and 512 $\times$ 512 during training. However, since RSIVQA comprises multiple datasets with varying image sizes, we first resize all images to a unified size of 256 $\times$ 256 before feeding them into the model.

\subsection{Performance Comparison}

In this section, we make a comparison of the proposed method against the current five RS-VQA approaches, listed as follows. Comparative studies are employed on RS-VQA LR, RS-VQA HR, and RS-IVQA datasets.

\subsubsection{\textbf{RSVQA}}

This approach fuses visual and textual features using dot products to predict answers \cite{lobry2020rsvqa}.

\subsubsection{\textbf{EasyToHard}} 

A text-guided multi-level visual feature learning approach, which incorporates the SPCL learning strategy. This approach gradually increases the difficulty of the QA pairs, starting from easier questions and progressing to more challenging ones \cite{yuan2022easy}.

\subsubsection{\textbf{MAIN}}

A model that leverages attention mechanisms and bilinear techniques to enhance the relationship between spatial positions and textual information \cite{zheng2021mutual}.

\subsubsection{\textbf{Bi-Modal}} 

A model based on an encoder-decoder architecture that utilizes co-attention to obtain integrated features for answer prediction \cite{bazi2022bi}.

\subsubsection{\textbf{SHRNet}}
A model utilizes a hash-based spatial multiscale visual learning method to enhance the perception of spatial positional information, thereby obtaining improved image features for fusion \cite{zhang2023spatial}. 

\begin{table}
\centering
\caption{Comparison with the state of the art on the LR test set. All values are reported as percentage (\%), with the maximum value of each entry in bold.}
\label{tab:LR_1}
\resizebox{\linewidth}{!}{
\begin{tabular}{lcccccc}
\hline
\multirow{2}{*}{Model} & \multicolumn{4}{c}{Types} & \multirow{2}{*}{\makecell[c]{Average \\ Accuracy}}      & \multirow{2}{*}{\makecell[c]{Overall \\ Accuracy}} \\ \cline{2-5}    
& Count   & Presence & Comparison & Rural/Urban \\ \hline
RSVQA \cite{lobry2020rsvqa}      & 67.01 & 87.46  & 81.50    & 90.00     & 81.49 & 79.08 \\
EasyToHard \cite{yuan2022easy} & 69.22 & 90.66  & 87.49    & 91.67     & 84.76 & 83.09 \\
Bi-Modal \cite{bazi2022bi}    & 72.22 & 91.06  & 91.16    & 92.66     & 86.78 & 85.56 \\
SHRNet \cite{zhang2023spatial}     & 73.87 & 91.03  & 90.48    & \textbf{94.00}     & 87.34 & 85.85 \\ \hline
Ours(question-only) & 61.03 & 89.83 & 89.55 & 55.00 & 73.85 & 80.90 \\
Ours    & \textbf{75.07}    & \textbf{92.29}    & \textbf{92.10}    &91.67  & \textbf{87.78}    & \textbf{87.14} \\ \hline
\end{tabular}
}
\end{table}

\begin{table}
\centering
\caption{Comparison with the state of the art on the HR test set 1. All values are reported as percentage (\%), with the maximum value of each entry in bold.}
\label{tab:HR_1}
\resizebox{\linewidth}{!}{
\begin{tabular}{lcccccc}
\hline
\multirow{2}{*}{Model} & \multicolumn{4}{c}{Types} & \multirow{2}{*}{\makecell[c]{Average \\ Accuracy}}      & \multirow{2}{*}{\makecell[c]{Overall \\ Accuracy}} \\ \cline{2-5}  
& Count   & Presence & Comparison & Area \\ \hline
RSVQA \cite{lobry2020rsvqa}           & 68.63          & 90.43          & 88.19    & 85.24 & 83.12 & 83.23 \\
EasyToHard \cite{yuan2022easy}      & 69.06          & 91.39         & 89.75    & 85.92 & 83.97 & 84.16 \\
Bi-Modal \cite{bazi2022bi}         & 69.80          & 92.03          & 91.83    & 86.27 & 84.98 & 85.30 \\
SHRNet \cite{zhang2023spatial}          & 70.04          & \textbf{92.45}          & 91.68  & 86.35 & 85.13 & 85.39 \\ \hline
Ours(question-only) &  65.81 & 88.41 & 85.57 & 79.33 & 79.78 & 80.23 \\
Ours       & \textbf{70.47} & 92.43  & \textbf{92.20}    & \textbf{86.99}    & \textbf{85.52} & \textbf{85.81} \\ \hline
\end{tabular}
}
\end{table}

\begin{table}
\centering
\caption{Comparison with the state of the art on the HR test set 2. All values are reported as percentage (\%), with the maximum value of each entry in bold.}
\label{tab:HR_2}
\resizebox{\linewidth}{!}{
\begin{tabular}{lcccccc}
\hline
\multirow{2}{*}{Model} & \multicolumn{4}{c}{Types} & \multirow{2}{*}{\makecell[c]{Average \\ Accuracy}}      & \multirow{2}{*}{\makecell[c]{Overall \\ Accuracy}} \\ \cline{2-5}  
& Count   & Presence & Comparison & Area \\ \hline
RSVQA \cite{lobry2020rsvqa}       & 61.47 & 86.26    & 85.94      & 76.33 & 77.50 & 78.23 \\
EasyToHard \cite{yuan2022easy}  & 61.95 & 87.97    & 87.68      & 78.62 & 79.06 & 79.29 \\
Bi-Modal \cite{bazi2022bi}    & 63.06 & 89.37    & \textbf{89.62}      & 80.12 & 80.54 & 81.23 \\
SHRNet \cite{zhang2023spatial}    & \textbf{63.42} & 89.81    & 89.44      & 80.37 & 80.76 & 81.37 \\ \hline
Ours(question-only) &  61.70 & 87.48 & 86.27 & 76.99 & 78.11 & 78.81 \\
Ours       & 63.23 & \textbf{90.22}   & 89.49      & \textbf{81.66} & \textbf{81.15} & \textbf{81.68} \\ \hline
\end{tabular}
}
\end{table}

\begin{table}
\centering
\caption{Comparison with the state of the art on the RSIVQA dataset. All values are reported as percentage (\%), with the maximum value of each entry in bold.}
\label{tab:RSIVQA_1}
\resizebox{0.9\linewidth}{!}{
\begin{tabular}{lccccc}
\hline
\multirow{2}{*}{Model} & \multicolumn{3}{c}{Types} & \multirow{2}{*}{\makecell[c]{Average \\ Accuracy}}    & \multirow{2}{*}{\makecell[c]{Overall \\ Accuracy}} \\ \cline{2-4} 
& Yes/No  & Number  & Others    \\ \hline
MAIN \cite{zheng2021mutual}       & 92.82 & 56.71 & 54.50  & 68.01 & 77.39 \\
EasyToHard \cite{yuan2022easy} & 95.49 & 49.03 & 63.65  & 69.39 & 79.70 \\
SHRNet \cite{zhang2023spatial}     & 97.64 & 57.89 & 84.60  & 80.04 & 84.46 \\ \hline
Ours(question-only) &  84.04 & 48.60 & 7.73 & 46.79 & 63.78 \\
Ours       & \textbf{97.90} & \textbf{62.64}    & \textbf{92.47} & \textbf{84.34}    & \textbf{87.10} \\ \hline
\end{tabular}
}
\end{table}

We present a performance comparison between our method and other models on the LR dataset in Table \ref{tab:LR_1}. Noticeably, our method outperforms all other approaches in the majority of evaluation metrics. Remarkably, RSAdapter achieves an Average Accuracy (AA) of 87.78\% and an Overall Accuracy (OA) of 87.14\%, demonstrating substantial improvements of 0.44\% and 1.29\%, respectively, when compared to the latest SHRNet model. Compared to the baseline RSVQA model, it achieves even more remarkable improvements of 6.29\% and 8.06\%, respectively.
Specifically, within the four distinct question categories, RSAdapter achieves the best results in Count, Presence, and Comparison, with improvements of around 1\% in each category. However, in the Rural/Urban category, our method falls short of SHRNet's performance. This discrepancy can be attributed to the fact that only 1\% of the dataset is allocated to this category, underscoring the challenge of achieving optimal results for sparsely represented categories in imbalanced datasets.

In Tables \ref{tab:HR_1} and \ref{tab:HR_2}, we present a separate comparison of results for all methods on two different test sets from the HR dataset using the same evaluation metrics. Our method has demonstrated improvements compared to other models on both test sets. On the first test set, we achieve an AA of 85.52\% and an OA of 85.81\%. On the second test set, our AA and OA are 81.15\% and 81.68\%, respectively. Our method also exhibits superior performance in the majority of categories, with only a small gap (0.02\%-0.19\%) compared to the best model in other categories. It is important to note that the performance on the second test set is significantly lower, with a decrease of 4.37\% for AA and 4.13\% for OA compared to the first test set. This discrepancy is primarily due to the fact that the image data in the second test set originates from a different region compared to the rest of the dataset and employs different sensors for capture. This underscores the significant challenge posed by the data-shift problem for current methods. Furthermore, the differences between various methods in the HR dataset are noticeably smaller compared to the LR dataset. This indicates that with an increase in training data, the disparities between the results of different methods gradually diminish.

In Table \ref{tab:RSIVQA_1}, we present a performance comparison between RSAdapter and three other models on the RSIVQA dataset. Our method has achieved an AA of 84.34\% and an OA of 87.10\% on the RSIVQA dataset, representing significant improvements (4.3\% increase for AA and 2.76\% increase for OA) compared to the SHRNet model.
In the 'Others' category, we achieve an accuracy of 92.47\%, representing a substantial improvement of 7.87\% compared to the best results from other models. In the 'Number' category, we also observe a notable improvement of 5.93\%. However, it's worth noting that the performance in the 'Number' category improved by only 0.26\% compared to SHRNet, indicating that extracting quantity-related features from images remains a challenging task for the current method.

In addition, we have separately reported the results of the question-only model in Table \ref{tab:LR_1} to Table \ref{tab:RSIVQA_1}. This model masks all input images as 0 to predict answers, demonstrating the performance of the model when only textual information is provided. The final results show that the question-only model exhibits lower accuracy compared to the image-text model.

Overall, our transformer-based approach has shown improvements compared to the previous traditional CNN+RNN methods on three different datasets. This indicates that transformer methods, which have already achieved significant success in traditional computer vision and natural language processing fields, can also be effectively applied in the remote sensing domain and get comparable performance.

\subsection{Ablation Studies}

We conduct a comprehensive set of experiments to assess the effectiveness of key components of our model on three distinct datasets. Furthermore, we conduct studies to analyze the influence of different backbone models, data efficiency strategies, various bottleneck dimension sizes, the placement of RSAdapter insertion, and the number of transformer layers on the outcomes. Unless otherwise specified, our experimental setup remained consistent with the default settings.

\begin{table}
\centering
\caption{Effectiveness of proposed components on three different data sets. All values are reported as percentage (\%).}
\label{tab:component}
\resizebox{\linewidth}{!}{
\begin{tabular}{llcccccccc}
\hline
Dataset & Methods & \makecell[c]{Param \\ (M)} & \makecell[c]{Tunable \\ Param (M)} & \makecell[c]{Average \\ Accuracy}      & \makecell[c]{Overall \\ Accuracy}     \\ \hline
\multirow{7}{*}{LR}&Linear probing  & 113   & 1.2   & 83.86 & 82.86 \\
&Full finetune   & 113   & 113   & 87.22 & 86.47 \\ \cline{2-6}
&Bitfit \cite{zaken2022bitfit}   & 113   & 1.3  & 86.35 & 84.72 \\
& Adapter \cite{houlsby2019parameter}   & 120   & 8.3   & 87.56 & 86.55 \\
&Lora \cite{hu2021lora}  & 120   & 8.3   & 87.16 & 86.37 \\ \cline{2-6}
&RSAdapter(MSA)    & 116   & 4.8   & 86.75 & 86.33\\
&RSAdapter(MLP)  & 116   & 4.8      & 87.63 & 86.73 \\
&Scaling RSAdapter    & 120   & 8.4  & \textbf{87.78} & \textbf{87.14} \\ \hline
\multirow{7}{*}{HR1}&Linear probing  & 113   & 1.3 & 82.29 & 82.43   \\
&Full finetune   & 113   & 113   & 85.31 & 85.64    \\ \cline{2-6}
&Bitfit \cite{zaken2022bitfit}   & 113   & 1.4  & 84.84 & 85.12   \\
& Adapter \cite{houlsby2019parameter} & 120   & 8.4   & 85.18 & 85.49 \\
&Lora \cite{hu2021lora}  & 120   & 8.3  & 84.68 & 84.94   \\ \cline{2-6}
&RSAdapter(MSA)    & 116   & 4.8  & 85.26 & 85.54   \\
&RSAdapter(MLP)  & 116   & 4.8 & 85.35 & 85.67 \\
&Scaling RSAdapter    & 120   & 8.4      & \textbf{85.52} & \textbf{85.81} \\ \hline
\multirow{7}{*}{HR2}&Linear probing  & 113   & 1.3 & 77.58 & 78.30   \\
&Full finetune   & 113   & 113  & 80.79 & 81.40    \\ \cline{2-6}
&Bitfit \cite{zaken2022bitfit}   & 113   & 1.4  & 79.68 & 80.42   \\
& Adapter \cite{houlsby2019parameter}   & 120   & 8.4   & 80.25 & 80.77    \\
&Lora \cite{hu2021lora}  & 120   & 8.3  & 80.05 & 80.70   \\ \cline{2-6}
&RSAdapter(MSA)    & 116   & 4.8  & 80.46 & 81.05    \\
&RSAdapter(MLP)  & 116   & 4.8 & 80.54 & 81.16       \\
&Scaling RSAdapter    & 120   & 8.4 & \textbf{81.15} & \textbf{81.68}     \\ \hline
\multirow{7}{*}{RSIVQA} &Linear probing  & 114   & 2.2   & 78.71 & 83.15 \\
&Full finetune   & 114   & 114   & 84.09 & 87.27    \\ \cline{2-6}
&Bitfit \cite{zaken2022bitfit}   & 114   & 2.3 & 83.06   & 86.61  \\
& Adapter \cite{houlsby2019parameter}   & 121   & 9.3   & 83.18 & 86.49 \\
&Lora \cite{hu2021lora}  & 121   & 9.3  & 82.18 & 85.57   \\ \cline{2-6}
&RSAdapter(MSA)   & 117   & 5.8 & 83.61 & 86.55 \\
&RSAdapter(MLP)  & 117   & 5.8   & 84.10  & 86.88 \\
&Scaling RSAdapter    & 121   & 9.4   & \textbf{84.34}  & \textbf{87.10} \\ \hline
\end{tabular}
}
\end{table}

\subsubsection{Effectiveness of Components}

In Table \ref{tab:component}, we present a detailed comparison of different components of RSAdapter and other PEFT methods, including Bitfit \cite{zaken2022bitfit}, Adapter \cite{houlsby2019parameter}, and Lora \cite{hu2021lora}. Additionally, we report the results of linear probing and full fine-tuning for reference. {The setup for linear probing is as follows: all parameters of the ViLT model are frozen, and only the parameters of the final classifier are updated.

To analyze the impact of different RSAdapter insertion positions, we provide two variants: RSAdapter(MSA), inserted in parallel alongside MSA, and RSAdapter(MLP), inserted in parallel alongside MLP. From the table, it is evident that the full RSAdapter consistently achieves the best results across all three different test sets. Notably, RSAdapter(MLP) exhibits varying degrees of improvement over RSAdapter(MSA) across all datasets. This suggests that during the model's efficient fine-tuning stage, tuning the MLP tends to be more beneficial than tuning the MSA. When compared to other PEFT methods, the full RSAdapter significantly outperforms them when tuning an equivalent number of parameters. It is worth noting that, in the majority of datasets, the results of the Adapter method are superior to the other two methods, further confirming the effectiveness of the Adapter approach. Compared to full fine-tuning of the entire model, our approach achieves comparable results with only a small fraction of parameters tuned and, in some datasets, even exhibits slight improvements.

\subsubsection{Impact of skip connection in RSAdapter}

Table \ref{tab:skip_connection} presents a comparison of the results of RSAdapter with skip connection removed and RSAdapter with skip connection retained in different scenarios. We observe that removing the skip connection consistently results in better AA and OA in all scenarios. It is worth noting that compared to RSAdapter with skip connection added only alongside MLP, adding skip connection alongside both MSA and MLP in RSAdapter simultaneously can lead to a decrease in performance. Therefore, we have decided to remove the skip connection in RSAdapter.

\begin{table}
\centering
\caption{Performance comparision between RSAdapter with and without skip connection on the LR test set. All values are reported as percentage (\%).}
\label{tab:skip_connection}
\resizebox{\linewidth}{!}{
\begin{tabular}{lcccc}
\hline
Type & \makecell[c]{Param \\ (M)} & \makecell[c]{Tunable \\ Param (M)} & \makecell[c]{Average \\ Accuracy}      & \makecell[c]{Overall \\ Accuracy}     \\ \hline
MSA w/ sc  & 116   & 4.8   & \textbf{86.95} & 85.94  \\
MSA w/o sc    & 116   & 4.8  & 86.75 & \textbf{86.33} \\ \hline
MLP w/ sc  & 116   & 4.8  & 87.07 & 86.47  \\
MLP w/o sc    & 116   & 4.8  & \textbf{87.63} & \textbf{86.73} \\ \hline
MSA+MLP w/ sc & 120   & 8.4  & 86.58 & 86.38  \\
MSA+MLP w/o sc    & 120   & 8.4  & \textbf{87.78} & \textbf{87.14} \\ \hline

\end{tabular}
}
\\
\vspace{2pt}
\leftline{\scriptsize Note: sc indicates skip connection.}
\end{table}

\subsubsection{Different Backbone Models} 

In further validation of the applicability of our method, we conduct additional experiments using two unimodal large models, ViT \cite{dosovitskiy2020image} and BERT \cite{devlin2018bert}. Since ViT and BERT can only handle one modality of input each, we perform element-wise multiplication between the image features and text features obtained from these two models. Subsequently, we feed this combined image-text feature into the classification layer to obtain our answers. From Table \ref{tab:ViT_BERT}, it is evident that our method, with only half the parameter updates compared to ViT+BERT, outperforms it on three different datasets when using the ViLT model. This suggests that, even with a smaller parameter count in the multimodal backbone model, our method performs better than using unimodal backbone models.

\begin{table}
\centering
\caption{Performance comparision using different Backbone on three different data sets. All values are reported as percentage (\%).}
\label{tab:ViT_BERT}
\resizebox{\linewidth}{!}{
\begin{tabular}{llcccccccc}
\hline
Dataset & Backbone & \makecell[c]{Param \\ (M)} & \makecell[c]{Tunable \\ Param (M)} & \makecell[c]{Average \\ Accuracy}      & \makecell[c]{Overall \\ Accuracy}     \\ \hline
\multirow{2}{*}{LR}& ViT-B+BERT-B  & 210   & 15.5   & 87.13 & 86.20 \\
&ViLT-B    & 120   & 8.4  & \textbf{87.78} & \textbf{87.14} \\ \hline
\multirow{2}{*}{HR1}& ViT-B+BERT-B  & 210   & 15.5  & 85.22 & 85.58   \\
&ViLT-B    & 120   & 8.4  & \textbf{85.52} & \textbf{85.81} \\ \hline
\multirow{2}{*}{HR2}& ViT-B+BERT-B  & 210   & 15.5  & 80.83 & 81.50  \\
&ViLT-B    & 120   & 8.4  & \textbf{81.15} & \textbf{81.68} \\ \hline
\multirow{2}{*}{RSIVQA}& ViT-B+BERT-B   & 211   & 16.5   & 83.89 & 86.64   \\
&ViLT-B    & 121   & 9.4  & \textbf{84.34} & \textbf{87.10} \\ \hline

\end{tabular}
}
\end{table}

\subsubsection{Data Efficiency}

\begin{table}
\centering
\caption{Data efficiency results on three different data sets. Results are reported as percentage (\%). Best results are highlighted in bold.}
\label{tab:data_effiency}
\resizebox{0.6\linewidth}{!}{
\begin{tabular}{lccc}
\hline
Dataset & Size    & \makecell[c]{Average \\ Accuracy}      & \makecell[c]{Overall \\ Accuracy}     \\ \hline
\multirow{4}{*}{LR}  & 0.1  & 83.70  & 83.35 \\
&0.2 & 85.96 & 84.70  \\
&0.5 & 87.37 & 85.55 \\
&1 & \textbf{87.78}    & \textbf{87.14}\\ \hline
\multirow{4}{*}{HR1}  & 0.1 & 84.08 & 84.32 \\
& 0.2 & 84.62 & 84.89 \\
& 0.5 & 85.27   & 85.54 \\
&1 & \textbf{85.52}    & \textbf{85.81}\\ \hline
\multirow{4}{*}{HR2}  & 0.1    & 79.50 & 80.30  \\
& 0.2   & 80.15 & 80.90 \\
& 0.5   & 80.81 & 81.50 \\
&1  & \textbf{81.15} & \textbf{81.68} \\ \hline
\multirow{4}{*}{RSIVQA} &0.1  & 75.12  & 81.02 \\
&0.2 & 78.47 & 82.98   \\
&0.5 & 82.23 & 85.67  \\
&1 & \textbf{84.34}    & \textbf{87.10}\\ \hline
\end{tabular}
}
\end{table}

To assess the data efficiency of our method, we analyze RSAdapter's performance under conditions of limited training data. We employ dataset subsets representing proportions of 0.1, 0.2, and 0.5 and report the results of all three datasets in Table \ref{tab:data_effiency}. Remarkably, even with very limited data, we are able to achieve acceptable results. Notably, our method can outperform other models trained on the complete dataset, even when using only half of the data (e.g., achieving a 2.19\% increase for AA and a 1.21\% increase for OA on the RSIVQA dataset). In Fig. \ref{fig:3}, we compare the results of RSAdapter and Bi-Modal when using only 10\% and 20\% of the data on LR and HR datasets. RSAdapter outperforms Bi-Modal on both LR and HR datasets, except for the AA on 10\% of LR. This discrepancy is due to the class imbalance in the LR dataset, where the Rural/Urban category has a significant impact on the final AA. In Fig. \ref{fig:4}, we also present the results of RSAdapter on the RSIVQA dataset across various training subsets, different categories, AA and OA. In conclusion, our method demonstrates the ability to achieve satisfactory results even with limited data.

\begin{figure}
    \centering
    \includegraphics[width=\linewidth]{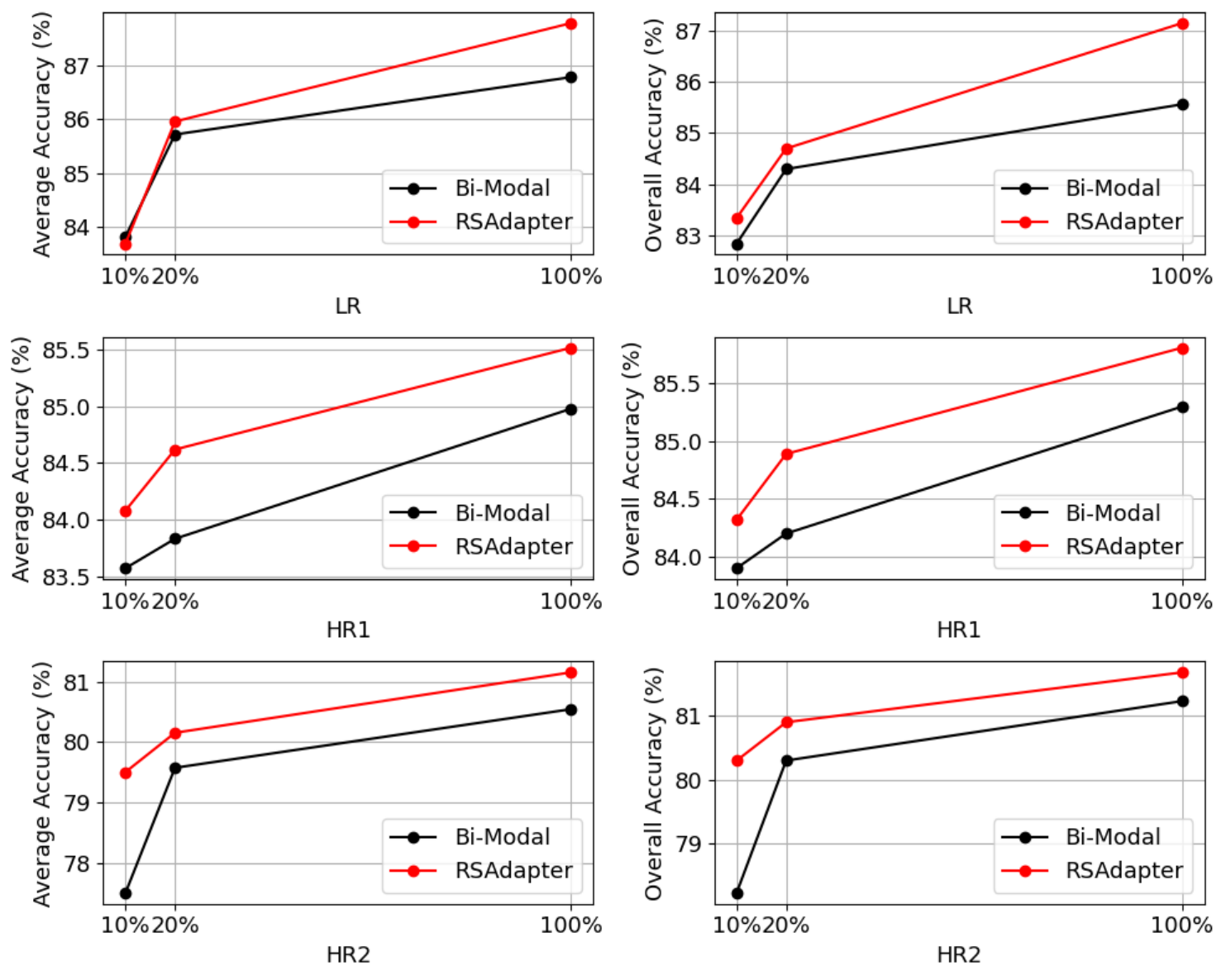}
    \caption{Data efficiency comparison with Bi-Modal on LR and HR datasets.}
    \label{fig:3}
\end{figure}

\begin{figure}
    \centering
    \includegraphics[width=\linewidth]{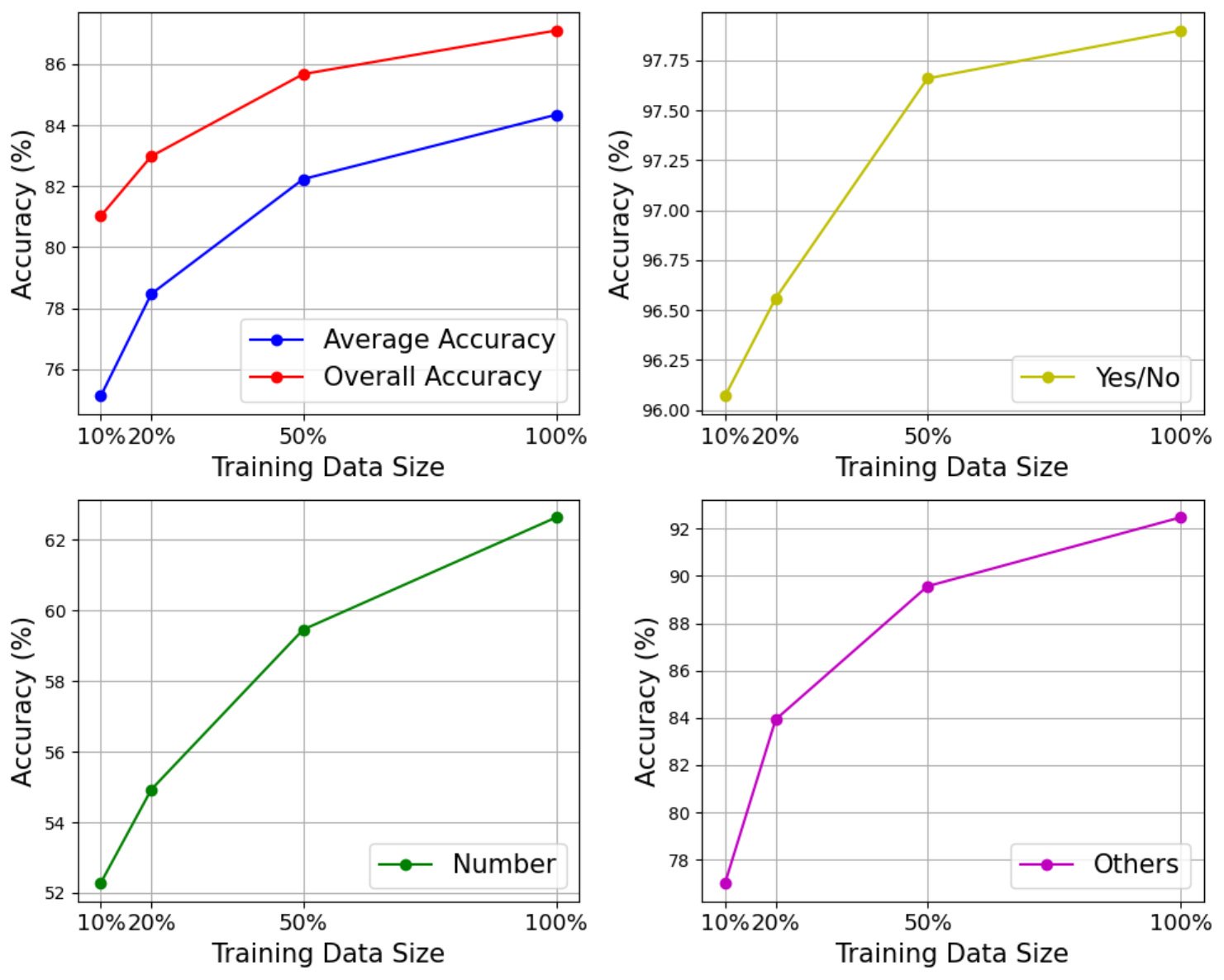}
    \caption{Data efficiency performance on RSIVQA dataset.}
    \label{fig:4}
\end{figure}

\subsubsection{Different Bottleneck Dimension Size}

\begin{table}
\centering
\caption{Effect of bottleneck size $d^{\prime}$ of RSAdapters on the LR test set. Results are reported as percentage (\%). Best results are highlighted in bold.}
\label{tab:LR_3}
\resizebox{0.8\linewidth}{!}{
\begin{tabular}{ccccccccc}
\hline
$d^{\prime}$    & \makecell[c]{Param \\ (M)} & \makecell[c]{Tunable \\ Param (M)} & \makecell[c]{Average \\ Accuracy}      & \makecell[c]{Overall \\ Accuracy}     \\ \hline
32  & 114   & 2.5   & 86.85 & 86.05   \\
64  & 115   & 3.7  & 87.44 & 86.53   \\
128 & 118   & 6.0   & 87.60 & 86.15   \\
192 & 120   & 8.4  & \textbf{87.78}    & \textbf{87.14}    \\
256 & 122   & 10.7  & 87.72 & 86.63     \\
384 & 127   & 15.5  & 86.96 & 86.29   \\ \hline
\end{tabular}
}
\end{table}

The bottleneck dimension size of RSAdapters is a crucial hyperparameter that affects both the number of model parameters and tunable parameters. To investigate its impact on model performance, we select $d^{\prime}$ from the set ${32, 64, 128, 192, 256, 384}$. Table \ref{tab:LR_3} presents the results of the model on the LR test set for different values of $d^{\prime}$. Upon analyzing the results, we observe that the model performs best when $d^{\prime}$ is set to 192. However, as $d^{\prime}$ increases, the model's parameter count also rises, but there is no corresponding improvement in performance. This suggests that the increase in $d^{\prime}$ may lead to increased optimization complexity, resulting in worse performance. Therefore, selecting the appropriate $d^{\prime}$ is crucial.

\subsubsection{Different Position of RSAdapters}

In addition to the experiments mentioned above, we also conduct experiments to assess the impact of the insertion positions of RSAdapter, using the same tunable parameters for a fair comparison. Table \ref{tab:LR_4} provides results for different insertion positions: 'Top' refers to inserting RSAdapter into the last 6 layers of the transformer, 'Bottom' refers to inserting RSAdapter into the first 6 layers of the transformer, while 'Even' and 'Odd' stand for inserting RSAdapter into the even and odd layers of the transformer, respectively. Remarkably, when RSAdapter is inserted into the lower layers, the results consistently outperform those when inserted into the upper layers. This could be attributed to the fact that there may be slight differences in low-level features between natural images and remote sensing images, leading to significant variations in results when fine-tuning a model pre-trained on natural images for remote sensing images.

\begin{table}
\centering
\caption{Effect of different positions of RSAdapters on the LR test set. Results are reported as percentage (\%). Best results are highlighted in bold.}
\label{tab:LR_4}
\resizebox{0.8\linewidth}{!}{
\begin{tabular}{lcccccccc}
\hline
Position    & \makecell[c]{Param \\ (M)} & \makecell[c]{Tunable \\ Param (M)} & \makecell[c]{Average \\ Accuracy}      & \makecell[c]{Overall \\ Accuracy}     \\ \hline
Top  & 116   & 4.8    & 86.84    & 86.12      \\
Bottom  & 116   & 4.8 & 87.75 & 86.67     \\
Even & 116   & 4.8    & 87.03 & 86.06      \\
Odd & 116   & 4.8 & 87.46 & 86.62      \\
All & 120   & 8.4  & \textbf{87.78}    & \textbf{87.14}   \\ \hline
\end{tabular}
}
\end{table}

\subsubsection{Different Number of Transformer Layers}

Finally, to verify the effectiveness of the model under varying computational resources, we conduct experiments with different numbers of transformer layers, as shown in Table \ref{tab:LR_5}. The results demonstrate that even with only 6 transformer layers, we can still achieve superior performance compared to all previous models. This reaffirms the effectiveness of our method.

\begin{table}
\centering
\caption{Effect of different number of transformer layers on the LR test set. Results are reported as percentage (\%). Best results are highlighted in bold.}
\label{tab:LR_5}
\resizebox{0.8\linewidth}{!}{
\begin{tabular}{ccccccccc}
\hline
\makecell[c]{Layer \\ Size}    & \makecell[c]{Param \\ (M)} & \makecell[c]{Tunable \\ Param (M)} & \makecell[c]{Average \\ Accuracy}      & \makecell[c]{Overall \\ Accuracy}     \\ \hline
3   & 50.8  & 3.0   & 85.09 & 84.45  \\
6   & 73.9  & 4.8   & 87.23 & 86.02      \\
9   & 96.9  & 6.6   & 87.54 & 86.43     \\
12  & 120   & 8.4   & \textbf{87.78}    & \textbf{87.14}   \\ \hline
\end{tabular}
}
\end{table}

\subsection{Language Bias}

For current RS VQA models, a significant concern is language bias. To verify whether our approach learns strong language biases, we conduct experiments on the RSIVQA dataset. Firstly, we define three scenarios: in Scenario 1, the input during training consists of original image-question pairs; in Scenario 2, the input remains the same, but during testing, the input image is replaced with a random test image, implying a high probability of mismatch between the image and the question. In the final Scenario 3, we train the model using questions and random images to assess its generalization ability. Table \ref{tab:LR_6} presents the experimental results. Compared to Full finetune, RSAdapter exhibits more decrease in accuracy when the test images are replaced. This indicates that our method learns more features from the images and less language bias. This result is also evident from the results of Scenario 3. The accuracy of our method is lower than that of Full finetune, suggesting that in the same circumstances, Full finetune relies more on questions and learns more language bias. Therefore, compared to Full finetune, our model exhibits stronger generalization and learns less language bias.

\begin{table}
\centering
\caption{Comparison of the performance of Full finetune and RSAdapter on the RSIVQA dataset in three different scenarios. All values are reported as percentage (\%).}
\label{tab:LR_6}
\resizebox{\linewidth}{!}{
\begin{tabular}{lccccc}
\hline
\multirow{2}{*}{Methods} & \multirow{2}{*}{Scenario} & \multicolumn{3}{c}{Types}   & \multirow{2}{*}{\makecell[c]{Overall \\ Accuracy}} \\ \cline{3-5} 
& & Yes/No  & Number  & Others    \\ \hline  
\multirow{3}{*}{Full finetune} & 1 & 97.94 & 64.24 & 90.10 & 87.27 \\
& 2 & 78.16 & 47.53 & 4.34 & 59.61 \\
& 3 & 84.04 & 49.01 & 7.80 & 63.91  \\ \hline
\multirow{3}{*}{RSAdapter}  & 1 & 97.90 & 62.64    & 92.47 & 87.10 \\
& 2 & 70.73 & 42.90 & 5.22 & 54.08 \\ 
& 3 & 83.69 & 48.69 & 7.73 & 63.60 \\ \hline
\end{tabular}
}
\\
\begin{flushleft}
\scriptsize Note: In Scenario 1, the input consists of the original images and questions from the dataset. In Scenario 2, the input is the same as Scenario 1, but random images are used during testing. In Scenario 3, the input consists of questions and random images.
\end{flushleft}
\end{table}

\subsection{Visulization}

\begin{figure*}
    \centering
    \includegraphics[width=0.87\textwidth]{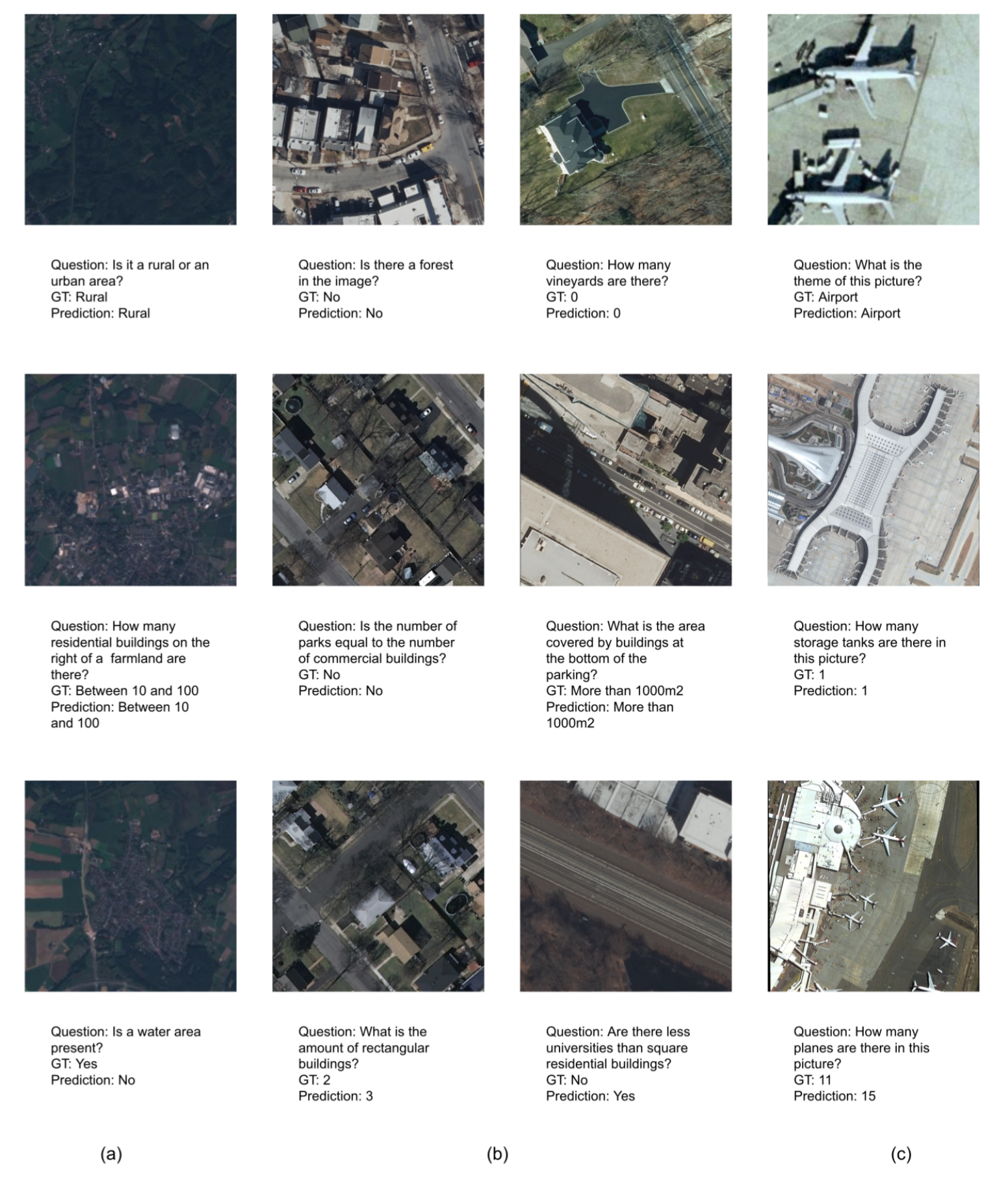}
    \caption{Typical remote sensing visual question answering examples on LR (a), HR (b) and RSIVQA (c) datasets.}
    \label{fig:5}
\end{figure*}

In this section, we begin by visualizing examples that compare the model's predictions with ground truth (GT) in Fig. \ref{fig:5}. The first and second rows in the images are examples where the answers are correctly predicted, while the bottom row represents examples where the answers are incorrectly predicted.

\begin{figure}
    \centering
    \includegraphics[width=\linewidth]{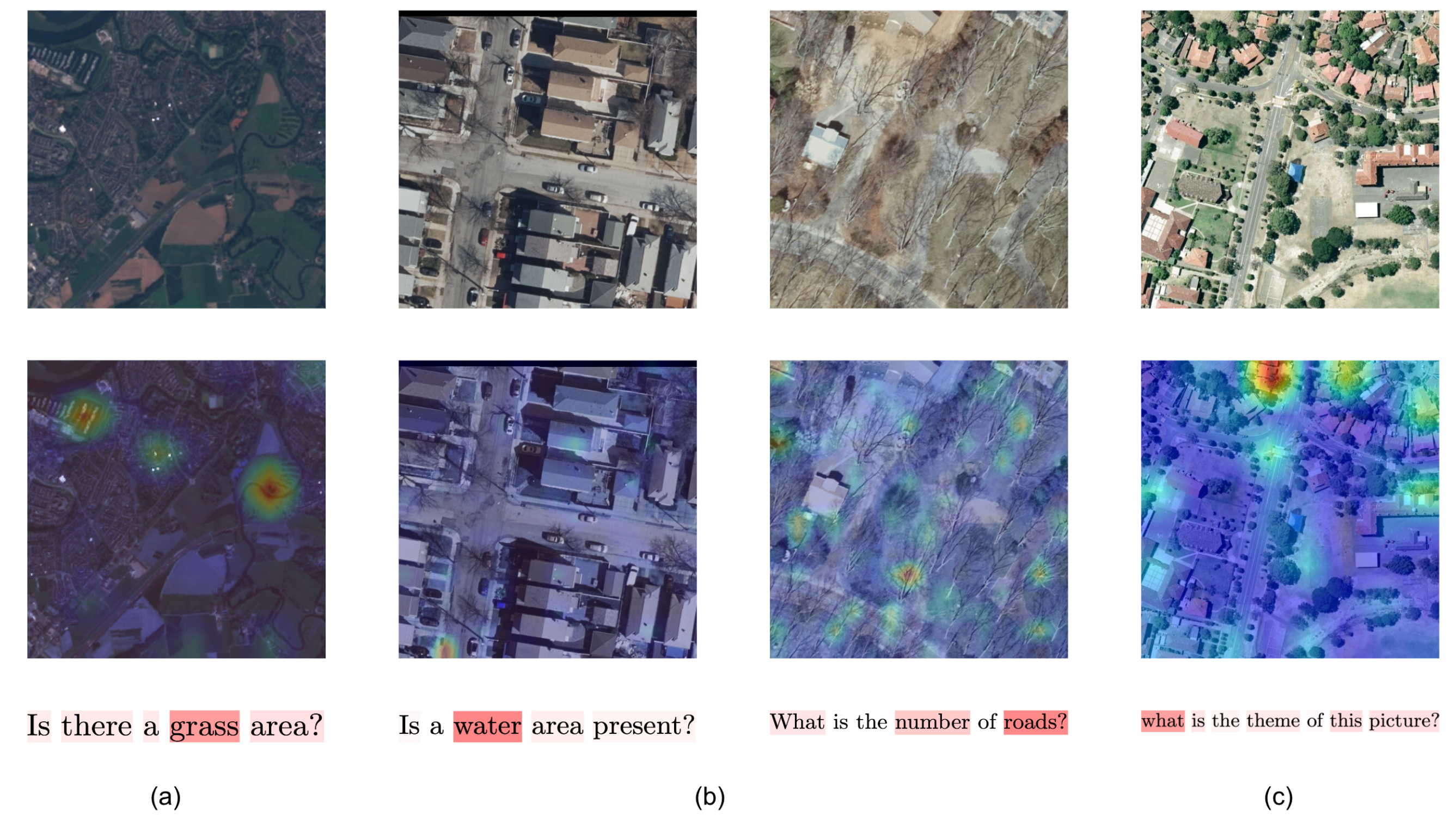}
    \caption{Typical remote sensing visual question answering examples on LR (a), HR (b) and RSIVQA (c) datasets. From top to bottom, the figure includes the original image, attention map over image, and attention map over text.}
    \label{fig:6}
\end{figure}

Then we demonstrate the attention map visualization of the RSAdapter over images and questions in Fig. \ref{fig:6}. We start by extracting the class token output from the last transformer layer along with the attention matrices for both image and text. We then average the results from all attention heads to obtain the final attention maps.

In Fig. \ref{fig:6}(a), we present an example from the LR dataset, which asks if there is a grass area in the image. Consequently, the model focuses primarily on the grass object. Similarly, for questions from the HR dataset, we showcase two examples in Fig. \ref{fig:6}(b). For the first example, this question asks if there is a water area in the image. Since there is no water present in the image, only a small portion of the image is attended to, indicating the absence of a water area in the image. For another HR example, the question involves counting the number of roads. Therefore, the model assigns varying levels of attention to objects resembling roads in order to infer the answer. In Fig. \ref{fig:6}(c), we also present an example from the RSIVQA dataset where the question pertains to the subject of the image. As a result, the model focuses on different objects. We can observe from attention map that the model successfully emphasizes residential objects.

While the examples presented demonstrate that the model effectively captures the relationship between images and text, it is worth noting that the model may still struggle with object recognition because specific object boundaries were not provided during training. Additionally, since many questions in the datasets are auto-generated, they might not always be relevant to the images, making it challenging to align the image and text relationships, which can affect the final performance of the model.

\subsection{Complexity Analysis}

\begin{table}
\centering
\caption{Comparison of complexity with other methods on LR dataset.}
\label{tab:LR_7}
\resizebox{\linewidth}{!}{
\begin{tabular}{lccccc}
\hline
Methods    & \makecell[c]{Param \\ (M)} & \makecell[c]{Tunable \\ Param (M)}    & \makecell[c]{Training \\ Time (s)} & \makecell[c]{Testing \\ Time (s)} & \makecell[c]{Overall \\ Accuracy(\%)} \\ \hline
RSVQA \cite{lobry2020rsvqa} & 85.7   & 5.7   & 154   & 28.04 & 79.08 \\
Bitfit \cite{zaken2022bitfit}   & 113   & 1.3  & 158 & 15.63 & 84.72 \\
Lora \cite{hu2021lora}  & 120   & 8.3   & 157 & 15.80 & 86.37 \\
Adapter \cite{houlsby2019parameter}   & 120   & 8.4   & 159 & 15.97 & 86.55 \\ \hline
RSAdapter w/o rep & 120   & 8.4 & 159   & 16.59 & 87.14  \\
RSAdapter w/ rep  & 120   & 8.4 & 159   & 16.03 & 87.14  \\ \hline
\end{tabular}
}
\end{table}

Here, we evaluate the complexity of RSAdapter. We assume that the model's feature dimension is represented as $d$, and the bottleneck size of RSAdapter is denoted as $d^{\prime}$. Consequently, the tunable parameters for each transformer layer during the training phase can be calculated as $2(d d^{\prime}) + 2(d + d^{\prime})$. During inference, the number of extra parameters is simplified to $2d d^{\prime}$.

Furthermore, we conduct a series of complexity comparisons with RSVQA \cite{lobry2020rsvqa}, and the results are presented in Table \ref{tab:LR_7}. The training and testing times represent the average time for one epoch, and the experiment was conducted on an Nvidia RTX A6000 GPU. From the results, we can infer that our method requires a similar amount of time for training one epoch compared to the RSVQA model, even though it has slightly more tunable parameters. However, our method achieves nearly twice the inference speed of RSVQA. Notably, SHRNet's training time is 3.8 times longer than RSVQA, and its testing time is 3.3 times longer than RSVQA based on \cite{zhang2023spatial}. This strongly suggests that our model provides a substantial improvement in runtime efficiency compared to SHRNet. In terms of parameter comparison, SHRNet has 105.56 million training parameters, whereas we achieve better results on all three datasets with only 10\% of SHRNet's parameter count. In addition, we compare our approach with other common PEFT methods. Compared to Bitfit and Lora, RSAdapter requires longer training and testing times due to the introduction of new modules by the Adapter. However, compared to Adapter, using RSAdapter results in a significant improvement in OA within the same training duration. Additionally, the model's inference speed increased by 3.38\% after applying re-parameterization techniques. These experiments demonstrate the runtime and parameter efficiency of our method.

\subsection{Application in RS}

\subsubsection{Scale Variation}

Images captured in remote sensing applications can vary significantly in scale, from large-scale satellite images to close-up aerial imagery. Addressing scale variation in image-text models is crucial for accurately understanding and interpreting remote sensing data.

\subsubsection{Dataset Diversity}

Remote sensing datasets often encompass diverse environments, terrains, and conditions. Ensuring that image-text models are trained on diverse datasets representative of various remote sensing scenarios is essential for robust performance across different contexts.

\subsubsection{Language Biases}

Language biases in RS VQA datasets can skew model performance and limit generalization, particularly in remote sensing applications. Mitigating language biases through balanced dataset curation and careful model training is essential.

\subsubsection{Model Interpretability}

Interpretable models are crucial in remote sensing applications, where decisions based on model outputs may have significant real-world consequences. Ensuring that image-text models provide interpretable explanations for their predictions can enhance trust and facilitate decision-making in remote sensing tasks.

\section{Conclusions}

In this paper,  we propose a novel approach, namely RSAdapter, to efficiently fine-tune pre-trained multimodal models, enabling them to better adapt to RS-VQA tasks. First, a novel re-parameterization method is applied to the parallel adapter, then we simultaneously insert RSAdapter next to MSA and MLP components while adding corresponding scaling layers to control the contribution of each RSAdapter to the model effectively. By conducting validation on three different RS-VQA datasets, we have achieved better performance compared to previous works. At the same time, our training parameters and inference time have significantly decreased. Furthermore, our method can be easily applied to other large models. Despite the numerous advantages of our method, it still has several limitations. The first limitation is that the current RSAdapter fine-tuning method still requires a substantial number of tunable parameters. The second limitation is that the current pre-trained multi-modal model is trained on natural images, and there may still be a gap when transferring to remote sensing images.

In future work, we aim to achieve efficient fine-tuning by pre-training multimodal models on a large number of remote sensing images, thereby avoiding the loss associated with transferring from natural images to remote sensing images. Additionally, we hope to discover better methods to further reduce the size of the training parameters without sacrificing performance as much as possible. Current RS-VQA datasets are designed based on templates, limiting the assessment of current algorithms to simpler questions and their reasoning abilities. A valuable research direction would be to create more complex benchmark datasets for visual question answering in the remote sensing community.


\ifCLASSOPTIONcaptionsoff
  \newpage
\fi

\bibliographystyle{IEEEtran}

\bibliography{RSVQA}

\end{document}